\documentclass[journal]{IEEEtran}
\usepackage[utf8]{inputenc}
\usepackage[fleqn]{amsmath}
\usepackage{graphicx}
\usepackage[ruled]{algorithm}
\usepackage[noend]{algpseudocode}
\usepackage{latexsym}
\usepackage{cite}

\ifCLASSINFOpdf
  \DeclareGraphicsExtensions{.pdf,.jpeg,.png}
\else
\fi
\ifCLASSOPTIONcompsoc
 \usepackage[caption=false,font=normalsize,labelfont=sf,textfont=sf]{subfig}
\else
  \usepackage[caption=false,font=footnotesize]{subfig}
\fi
\begin{document}
%
\title{Energy-Efficient Object Detection using Semantic Decomposition}
%
%
%

\author{Priyadarshini~Panda, \textit{ Student Member}, \textit{IEEE},
         Swagath~Venkataramani, Abhronil~Sengupta,\textit{ Student Member}, \textit{IEEE}, Anand~Raghunathan,~\IEEEmembership{Fellow,~IEEE}
        and Kaushik~Roy,~\IEEEmembership{Fellow,~IEEE}
\thanks{P. Panda, S.Venkataramani, A. Sengupta, A. Raghunathan and K. Roy are with the Department
of Electrical and Computer Engineering, Purdue University, West Lafayette,
IN, 47906 USA e-mail: (pandap, venkata0, asengup, raghunathan, kaushik)@purdue.edu.}
}

%
%

\markboth{Journal of \LaTeX\ Class Files,~Vol.~14, No.~8, August~2015}%
{Shell \MakeLowercase{\textit{et al.}}: Bare Demo of IEEEtran.cls for IEEE Journals}
%



\maketitle

\begin{abstract}
Machine-learning algorithms offer immense possibilities in the development of several cognitive applications. In fact, large scale machine-learning classifiers now represent the state-of-the-art in a wide range of object detection/classification problems. However, the network complexities of large-scale classifiers present them as one of the most challenging and energy intensive workloads across the computing spectrum. In this paper, we present a new approach to optimize energy efficiency of object detection tasks using semantic decomposition to build a hierarchical classification framework. We observe that certain semantic information like color/texture are common across various images in real-world datasets for object detection applications. We exploit these common semantic features to distinguish the objects of interest from the remaining inputs (non-objects of interest) in a dataset at a lower computational effort. We propose a 2-stage hierarchical classification framework, with increasing levels of complexity, wherein the first stage is trained to recognize the broad representative semantic features relevant to the object of interest. The first stage rejects the input instances that do not have the representative features and passes only the relevant instances to the second stage. Our methodology thus allows us to reject certain information at lower complexity and utilize the full computational effort of a network only on a smaller fraction of inputs to perform detection. We use color and texture as distinctive traits to carry out several experiments for object detection. Our experiments on the Caltech101/CIFAR10 dataset  show that the proposed method yields 1.93x/1.46x improvement in average energy, respectively, over the traditional single classifier model. 
\end{abstract}

%

\section{Introduction}
Object detection is one of the core areas of research in computer vision \cite{dubey2005recognition}. 
Machine-learning classifiers have proven to be very useful for implementing such detection tasks \cite{krizhevsky2012imagenet, simonyan2014very}. A detection task is basically a classification problem of distinguishing an object of interest from a host of input data. Traditionally, a single complex classifier model (shown in Fig. 1 (a)) is used to perform detection. Here, all the inputs are processed through the single model to detect the object of interest. However, in order to scale to more challenging object detection problems, the classifier models must become larger, which implies an increase in computational resuorces. With  computational efficiency becoming a primary concern across the computing spectrum, energy-efficient object detection is of great importance.


%
Interestingly, we note that in a real world dataset, a major portion of input images have some characteristic broad semantic features like color, texture etc. that are common to the object of interest. Consider the simple example of detecting a red Ferrari from a sample set of vehicle images consisting of motorbikes and cars. The first intuitive step is to recognize all red vehicles in the sample and then look for a Ferrari-shaped object from the sub-sample of red vehicles 
Thus, we can reduce and simplify the original sample set by utilizing the semantic information as we progress towards the primary object detection task. This simplification can potentially reduce the compute effort.
Based on this idea, we introduce semantic decomposition of inputs into characteristic broad features, like color (red) or shape (car) in the above example, and using the representative semantics to build a hierarchical classification framework, with increasing levels of complexity, for faster and more energy-efficient object detection.

%
\begin{figure}[t!]
\centering
\includegraphics[width=0.5\textwidth]{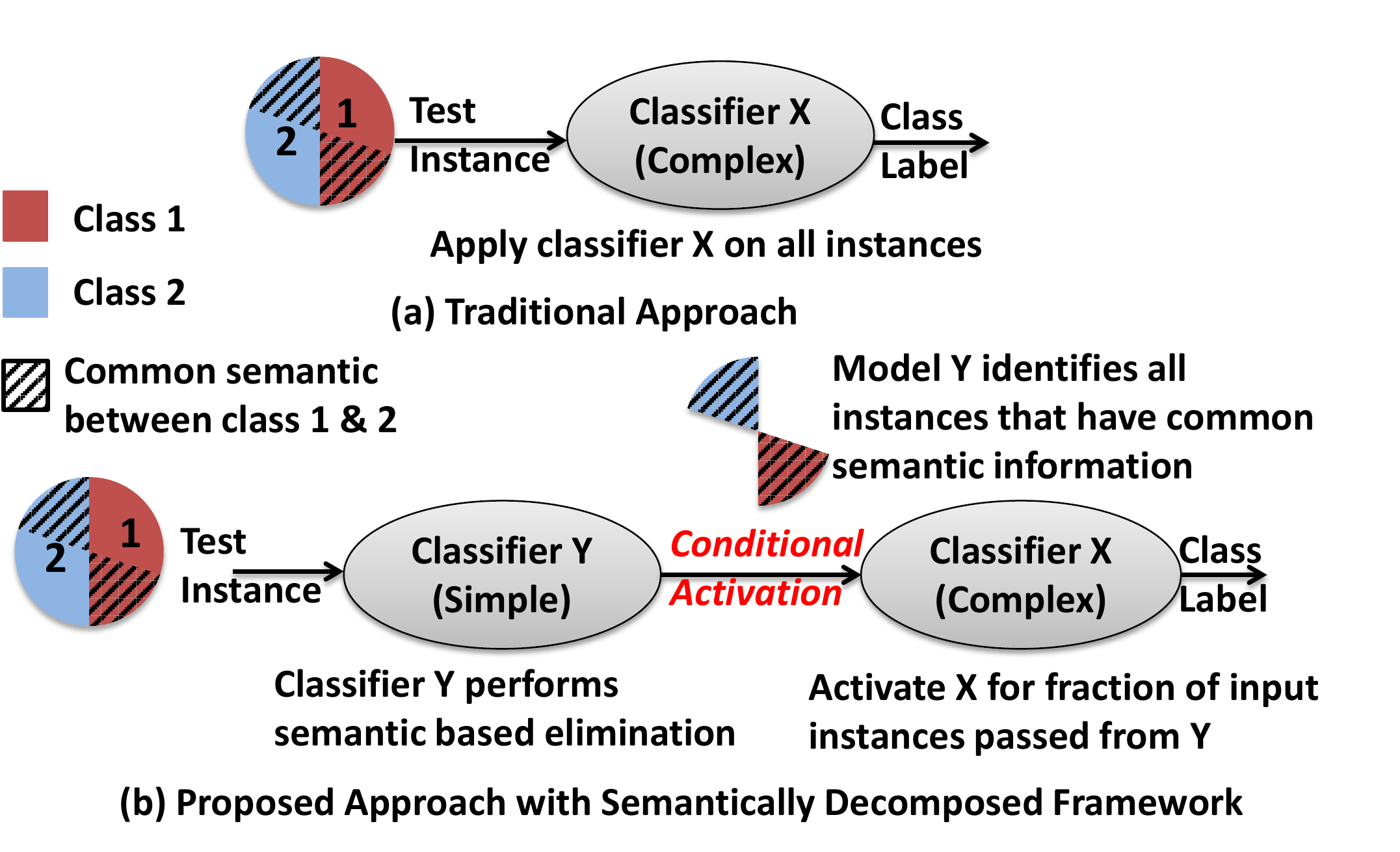}
\caption{(a) Traditional approach: learn one complex classifier and apply to all instances (b) Proposed approach: learn multiple simple classifiers on semantically decomposed features and activate complex model conditionally for instances that have common features with the object of interest. The simple classifiers in the first stage perform semantic based elimination.}
\vspace{-6mm}
\end{figure}

Fig. 1 illustrates our methodology. In the traditional approach shown in Fig. 1(a),
Classifier X is trained to separate data into two categories or classes: objects of interest that we are trying to detect (Class 2) and clutter or non-object of interest (Class 1). At test time, data instances are assigned to one class or the other depending on their labels. The computational effort in terms of energy and time to process every test instance depends on the complexity of the network i.e. the total number of weights and the neurons composing the classifier. In the example of Fig. 1(a), a single classifier clearly needs to be highly complex (more hidden neurons and hence, more synapses) in order to separate the classes with high accuracy. However, this leads to high computational effort for not only the test instances that have common semantic between the two classes but also the ones that do not share common features across the class labels.

In contrast, Fig. 1(b) shows our approach where we create a semantically decomposed framework with multiple classifiers (Y and X) with varying levels of complexity. Certain images in the dataset have a common semantic information representative of the object of interest shown by the shaded region in Fig. 1. Classifier Y is trained to identify all those instances that share the particular semantic with our object of interest (Class 2). It receives important yet simple semantically decomposed characteristics like color, edges, etc. from the input sensor data. The decomposed input features are simpler and easy to process than the original input image. Thus, the classifier in the first stage (Y) of the proposed framework are small scale (or less complex with few neurons and weights). The complex classifier X is then enabled for all those instances that have the semantic information that model Y is trained to detect. Hence, a significant potion of clutter (Class 1) are filtered out or eliminated at the first stage in this process. The second stage X, same as the complex classifier in the traditional approach, tries to detect the object of interest from the original input image. This approach can save time and energy, since all data instances need not be processed by the more complex traditional classifier. Please note that since the proposed methodology adds an extra classifier (first stage) into the overall classification framework, the additional cost overhead for the instances that are processed by both stages has to be taken into account in the computational cost. 

In order to observe maximum benefits and overcome the cost penalty (that the addition of first stage imposes), it is evident that the input dataset should have significantly larger clutter fraction than the objects of interest. Fortunately, in many useful detection applications, only a small fraction of the input dataset has relevant objects of interest. 
In \cite{venkataramani2015sapphire}, the authors have quantitatively established that in a wide range of video-based object detection datasets, only 5\% of the input data contains the relevant objects of interest. Our approach exploits this disproportionate distribution of input data to obtain compute-efficiency. The amount of time and energy saved also depends on how appropriately we decompose the input data or extract the common features such that we can reject as much clutter as possible in the first stage at very low complexity.

\section{Related Work}
On the algorithmic front, using multiple classifiers for increasing learning accuracy is an active area of research \cite{schapire2003boosting, deng2014ensemble, sun2013deep, gama2000cascade}. A class of work in ensemble based learning \cite{5444873,6626306,feitosa2014meta} exploit the idea that different classifiers can offer complementary information about patterns to be classified which can be used to improve the effectiveness of the overall recognition process. 
The main motivation behind such algorithmic techniques is to obtain an improvement in accuracy.  However, using them to reduce energy and runtime has received attention only in the recent past \cite{hosang2015makes}. The use of multiple classifiers in our proposed methodology is entirely driven by energy-efficiency and reduced computational complexity.

Past research in building energy-efficient neuromorphic systems have considered application-specific solutions \cite{vanhoucke2011improving,kaul20121}. In \cite{venkataramani2015scalable}, the authors have proposed a scalable effort classification framework consisting of a cascaded chain of biased classifiers with increasing complexity that can dynamically adjust the compute effort depending on the input data. The concensus between the classifier's outputs is used to decide if classification can be terminated at an earlier stage. The methodology that we propose in this paper is complementary to the concept of cascading classifiers. However, the novelty of our work arises from the fact that we leverage the semantic information in the input data to obtain efficiency and reduced testing complexity. Please note that though our training method draws inspiration from \cite{venkataramani2015scalable} , our method has different focus, design and evalutaion strategies.

Other popular approaches that have been explored to lower the compute effort of the network is based on approximate computing \cite{chippa2013energy,park2013saliency}. Exploiting the inherent error resilience of a system, a variety of approximate hardware \cite{venkataramani2013quality,chippa2014scalable} and software \cite{panda2015conditional,venkataramani2014axnn} techniques have been proposed to achieve reduced computational complexity. 
However, these techniques provide an explicit trade-off between efficiency and quality of results \cite{liu2009exploratory}. Our approach, on the other hand, provides energy savings, while maintaining classification accuracy.  



The rest of the paper is organized as follows. 
Section III describes our approach to construct the semantically decomposed framework. In Section IV, we discuss the methodology to train and test the semantic framework. In Section V and VI, we present the experimental methodology and results. We conclude in Section VII.

\section{Semantically Decomposed Object Detection}
In this section, we present our structured approach to design the hierarchical framework of classifiers. Optimum semantic selection and conditional activation of the classifier in the second stage form the bases of the framework. While there exists a suite of machine-learning classifiers like Support Vector Machines, Decision Trees , Neural Networks etc. suitable for object detection, we will focus on a particular class: Artificial Neural Networks (ANNs) to validate the proposed methodology for object detection. Please note that the semantically decomposed framework can be applied on other classification algorithms as well to lower the compute effort. In the rest of the paper, the terms 'Semantically decomposed framework' and 'Hierarchical framework' have been used interchangeably.

\subsection{Semantic Decomposition of Input Data}
In our proposed methodology, semantic decomposition is a very significant stage. In this stage, semantics such as texture or color components representative of the input image are extracted using appropriate image processing techniques. In this work, we use color and texture information individually in a set of experiments described in Section V as the first step of eliminating objects that do not share common semantic information. We use Hue-Saturation-Value (HSV) transformation \cite{levkowitz1993glhs} and Gabor filtering \cite{Jain97objectdetection} to extract color and texture components, respectively. Note that, after applying an HSV/Gabor transformation, an image in the HSV/Gabor space is much smaller as compared to the RGB space. The extracted feature vectors are then used as training instances to train the simpler (or less complex) classifiers in the first stage that filter out clutter from the objects of interest based on the absence of relevant semantic information. It is worth mentioning that the additional cost of HSV or Gabor processing also has to be taken into account for energy computations \cite{chen2008fast}. 

While we use color and texture as characteristic semantics, please note that other semantics like edges (with cany or sobel detectors), corners and blobs (with Laplacian of Gaussian) can also be used with the proposed methodology. In fact, for edge type attributes, feature descriptors like SIFT \cite{lowe2004distinctive} and HOG \cite{dalal2005histograms} can be used. For any of the feature descriptor used, it is important to include the additional cost for preprocessing while calculating the overall energy benefits. 

\subsection{Semantic based Elimination: Concept}
Fig. 2 (a) shows the conceptual view of the framework. All ANN models in both the traditional and hierarchical structure are learnt using the same algorithm and training data. In the traditional approach, as shown in Fig. 1(a), a single ANN model processes the RGB components of the image during training and classification.  In the 2-stage framework (Fig. 2(a)), each of the ANNs in the first stage are computationally efficient as they are trained on the optimal simple semantic feature vectors extracted from the original RGB image. The ANN in the second stage has a higher complexity on account of being fed the original RGB image for classification. Depending upon the output of the ANNs in the first stage, the second stage is enabled. This results in significant power savings due to conditional activation of the network in the second stage. The final classifier, same as the NN in the traditional structure, with the highest complexity makes sure that any clutter data that were passed onto the stage by the former ANNs due to misclassification are properly discarded or, classified as clutter in this stage, thereby maintaining the same classification accuracy as the traditional single classifier.
\begin{figure}[t!]
\centering
\includegraphics[width=1\linewidth]{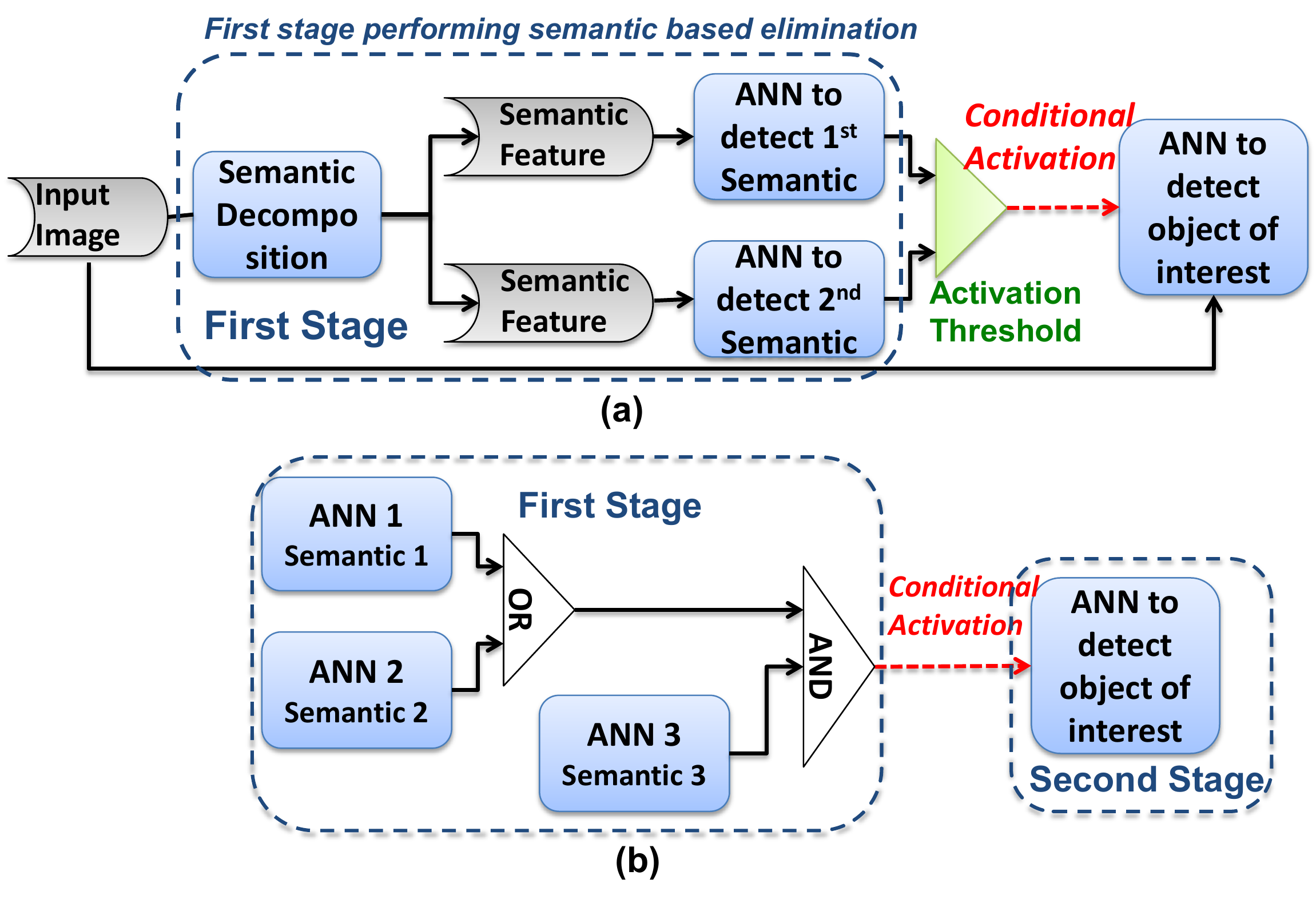}
\vspace{-7mm}
\caption{ (a) Proposed hierarchical  structure with  semantic decomposition  where multiple classifiers in the first stage detect the appropriate semantic followed by single complex classifier in the second stage enabled by the activation threshold (b) 2-level OR-AND first stage configuration with 2 optimal semantic features}
\vspace{-5mm}
\end{figure}
Besides the ANNs in the hierarchy, the setup also contains an activation threshold module (Fig. 2(a)). This module decides if the second stage should get enabled or not to determine the final output of the hierarchy for a given input image. Note that if the input is the desired object we are trying to detect, it will always be passed to the second stage. Then, the output of the hierarchy is based on the classification result of the second stage. Only when any clutter image is presented, the module then decides based on the following two criteria:
\begin{itemize}
\item If the classifiers in the first stage predict based on the semantic vector that the input presented is a clutter image, the second stage is not enabled and the output of the framework is a label corresponding to clutter.
\item If the classifiers in the first stage produce a sufficient confidence level on the input image’s semantic vector, the second stage is enabled. 
The framework’s final output is a label based on the result of the second stage.
\end{itemize}

\subsection{Optimal Semantic Selection}
The most important question that needs to be answered is how we select the appropriate semantics to construct the first stage configuration of the hierarchical structure. As mentioned earlier, addition of first stage imposes a cost penalty for inputs that need activation of both stages in the framework for correct classification (Fig. 2(a)). In contrast, in the traditional approach (Fig. 1(a)), only the single ANN needs to be enabled to get the final result. Thus, optimal semantic selection is key to obtaining a first stage with the lowest complexity so that the overall cost penalty can be reduced considerably.
To make our proposed approach more systematic, we devise an algorithm that recognizes the most optimum features and constructs a 2-level OR-AND configuration of the first stage for the most favorable classification results. Fig. 2(b) shows an example of the 2-level OR-AND first stage configuration for a given input dataset with 3 optimal semantic features. The given configuration implies that the second stage is only enabled when the first stage detects either Semantic 1 or 2 (OR) in combination with Semantic 3 (AND). In other words, the final ANN is enabled if ANN 3 and either of ANN 1 or 2 produce sufficient confidence level for a given input. 

\subsubsection{Efficiency and Accuracy Optimization with OR-AND configuration} 
To better understand the need for 2-level OR-AND configuration, consider the example shown in Fig. 3. Observe that the objects of interest can be characterized by two different semantic features. Hence, we need two different ANNs trained on two separate feature vectors in the first stage of the hierarchy. Fig. 3(a) shows that certain instances in the objects of interest have one semantic in common while the rest have the second semantic. So, we can choose the OR operation where the second stage/final classifier is enabled when we get a desired output from any one of the NNs in the first stage. In other words, final NN is enabled if one or the other semantic is identified from the input image. If the operation is set to be AND here, certain objects of interest will be rejected or misclassified in the first stage that will result in a significant decline in accuracy. On the other hand in Fig. 3(b), both semantics are present in all the instances of objects of interest.  While an OR would give a good result i.e. all objects of interest will be classified by the first stage and passed to the final classifier, however, the first stage would also pass a lot of clutter. This in turn would enable the final classifier for all the clutter passed resulting in a decline in efficiency. Thus, we need to set the activation as an AND operation where the final NN is enabled for inputs having both semantics i.e. we get a good confidence level for both the NNs in the initial stage. Fig. 3(b) clearly illustrates that the AND operation filters out a lot of clutter in the initial stage before forwarding to the final classifier. Thus, it is evident that while AND improves the efficiency of the classifier, OR increases the accuracy of the semantically decomposed framework. So, the semantic selection algorithm constructs the most optimal first stage OR-AND configuration that yields lowest computational cost with minimal loss in output quality in comparison to the baseline/traditional single classifier. 
\begin{figure}[t!]
\centering
\includegraphics[width=0.5\textwidth]{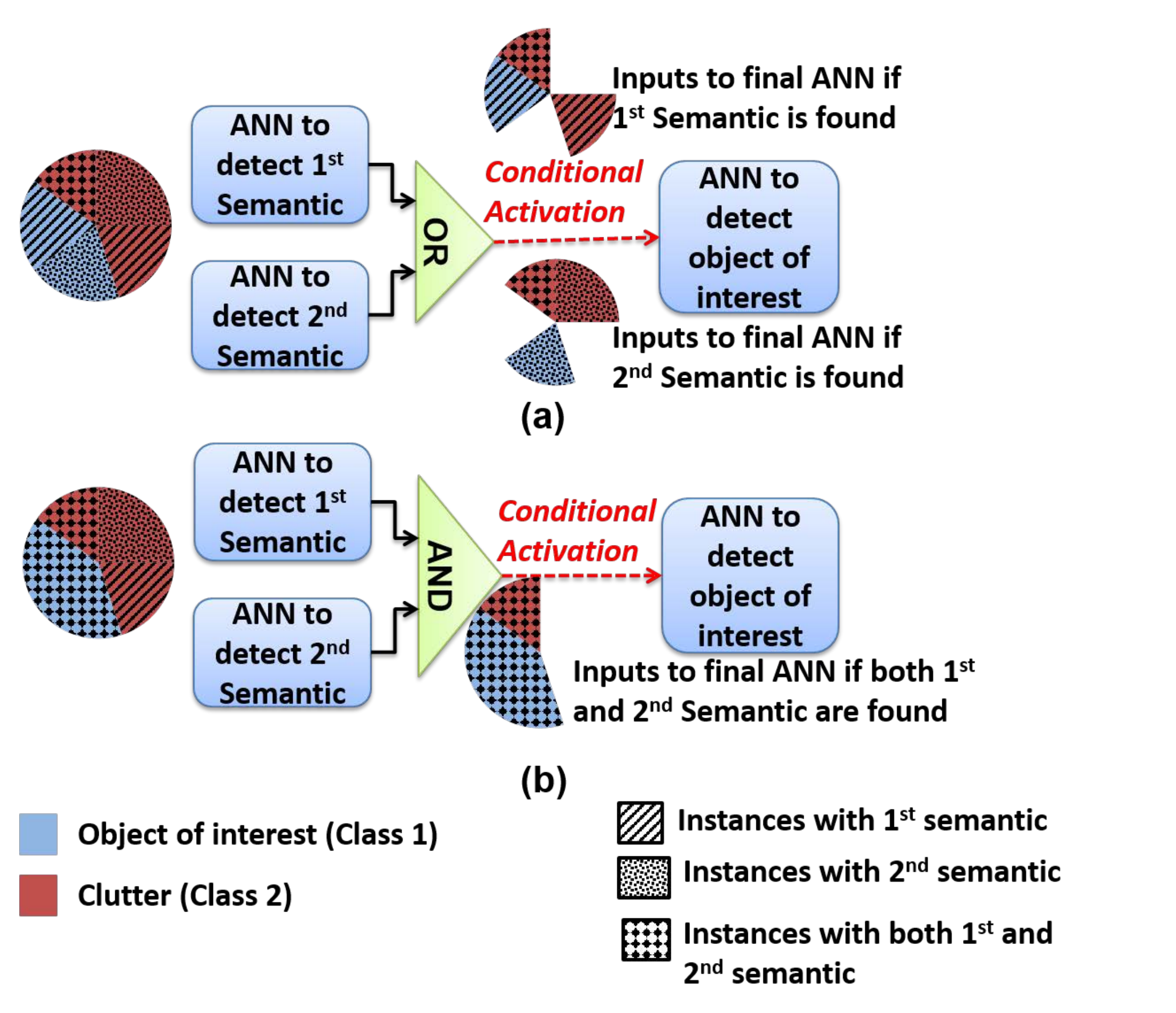}
\caption{(a) Two different semantic features not shared among all objects of Class 1 leading to OR operation for maximum accuracy (b) Two different semantic features common across all objects of Class 1 leading to AND operation for maximum accuracy as well as optimum efficiency}
\vspace{-5mm}
\end{figure}

\subsubsection{Adding ANNs or classifiers in first stage} 
While adding a new ANN to the first stage configuration, the algorithm calculates the overall gain obtained with conditional activation accounting for the additional penalty of the first stage. Consider a baseline ANN with computational cost $N_{orig}$ per instance. Let \textit{f} be the fraction of inputs that are filtered out or eliminated at the first stage. Let the computational cost of the first stage configuration of the semantically decomposed framework be $N_{initial}$. The following condition (Eq. 1) should be satisfied to improve the overall efficiency of the hierarchical framework:
\begin{equation}
N_{initial}+(1-f).N_{orig} < N_{orig}
\end{equation}
\begin{equation}
G= N_{orig} - [N_{initial}+(1-f).N_{orig}] 
\end{equation}

The left hand side of Eq. 1 represents the overall gain in efficiency with the semantically decomposed framework, which is the summation of first and second stage complexities. The second stage has a reduced complexity on account of conditional activation only for those fraction of inputs passed to that stage. This should be greater than the right side of Eq. 1 which denotes the original cost of the baseline/traditional single ANN classifier without any penalty from the addition of first stage. The difference of the left side from the right side is the overall gain (\textit{G}) as shown in Eq. 2.

\subsubsection{Semantic Selection Methodology} 
Algorithm 1 shows the pseudo code for selecting the optimal semantics and constructing the 2-level OR-AND configuration for the first stage of the semantically decomposed framework. The process takes the baseline/traditional single ANN $N_{orig}$,  training dataset $D_{tr}$ and the semantic feature search space as input and produces the optimal first stage $N_{initial}$ with appropriate OR-AND configuration. First, we train $N_{orig}$ on $D_{tr}$  and obtain the accuracy $Q$ (line 2). Next, we iteratively traverse through the semantic feature space selecting the feature (or combination of features) that improves the gain while maintaining the quality constraint (lines 3-20). The procedure terminates if adding a particular feature to the first stage doesn't improve the gain of the existing first stage configutation (line 8).
\setlength{\textfloatsep}{0pt}
\begin{algorithm}[t]
\caption{Pseudo code for optimal semantic selection and 2-level OR-AND construction of first stage}
\label{algo1}
 \textbf{Input:} Baseline classifier $N_{orig}$, training dataset $D_{tr}$, \# features in the search space N\\
 \textbf{Output:} First stage NN configuration: $N_{initial}$
\begin{algorithmic}[1]
\State \textbf{initialize} $N_{initial}$ = NULL, pairwise = FALSE, $G_0$= $\epsilon_1$
\State Train $N_{orig}$ using $D_{tr}$  and obtain the accuracy, $Q$. 
\State \textbf{for} $ i =1:N$ // \textit{for each feature vector in the search space}
\State Train a NN ($N_{i}$) on the feature vector \textit{i} using $D_{tr}$ and obtain accuracy, $Q'$ for the hierarchical framework with $N_{i}$ as first stage and $N_{orig}$ as second stage.
\State \textbf{if} ($Q-Q' < \epsilon$  // \textit{if quality constraint is met}
\State $N_{initial}'$ = $N_{i}$ AND $N_{initial}$, Calculate gain $G_i$ as per Eq. 8 with $N_{initial}'$ in the first stage and $N_{orig}$ as second stage.
\State \textbf{if} $G_i > G_{i-1}$ then  $N_{initial}$ =  $N_{initial}'$ // \textit{if gain improves then admit the new semantic $N_{i}$  ANDed with the existing first stage}
\State \textbf{elseif} ($G_i < G_{i-1}$ \&\& pairwise= TRUE) \textbf{TERMINATE} algorithm and return the existing first stage configuration $N_{initial}$ as output. // \textit{End the algorithm when there is no improvement in gain after exploring all combinations}
\State \textbf{end if}
\State \textbf{end if}
\State \textbf{end for} // \textit{by the end of the for loop, the first stage $N_{initial}$ is either NULL or an ANDed configuration of all optimal semantic NNs. Next, we look at pairwise combinations of features that can be ORed in the first stage. }
\State \textbf{remove} the semantic vectors already admitted into first stage $N_{initial}$ from the search space. \# features in the remaining search space$= N'$ , pairwise = TRUE, Select the top $k (k <N')$ features quality-wise from the $N'$ search space.
\State \textbf{for} $i=1:\binom{k}{2}$ // \textit{for each pair combination in the k-feature space}
\State $N_{initial}'$ = $N_1$ OR $N_2$ //\textit{where $N_1$ and $N_2$ are the two NNs corresponding to the semantic pair for the $i^{th}$ combination}
\State $N_{initial}^{temp}$  = $N_{initial}$ AND $N_{initial}'$ 
\State Obtain accuracy $Q'$ for the hierarchical framework with $N_{initial}^{temp}$ as first stage and $N_{orig}$ as second stage.
\State Repeat Steps 5-10 with $N_{i} \equiv N_{initial}^{temp}$  
\newline\textit{ if quality constraint is met and gain obtained with current configuration is higher than previous configuration, then,  \textbf{$N_{initial} = N_{initial}^{temp}$} }
\State \textbf{if}  $N_{initial}$ (current iteration) ==  $N_{initial}$ (previous iteration) then \textbf{continue}
\State \textbf{else} GOTO Step 12 and Repeat Steps 12-20
\State \textbf{end for}
\end{algorithmic}
\end{algorithm}

Initially, we search through the vector space and check the quality and gain constraints for each semantic vector (lines 3-11). The semantics that improve the overall gain, calculated as per Eq. 2 discussed earlier, are ANDed together and set as the initial stage ($N_{initial}$) (line 7). Next, we eliminate the semantic vectors already admitted into the initial stage and search through the remaining search space for pairwise ORed combinations of semantics from the top $k$ features that would improve the accuracy and the overall gain (lines 12-14). For datasets where inputs can be characterized by two different semantic features not shared among all the objects of interest (Fig. 3 (a)),  OR combination is essential to improving the accuracy of the hierarchical framework. It ensures that all objects of interest are passed to the second stage without being eliminated or misclassified by the first stage. The pairwise OR combination of NNs are then ANDed with the existing first stage (lines 15-17). If accuracy loss of the hierarchical framework with new first stage configuration with respect to the baseline is lesser than certain threshold $\epsilon$ (line 5), we check for the gain constraint. If the gain of the new configuration improves over the previous one, we select the corresponding semantic vectors and set the new OR-AND configuration as the first stage of the hierarchical framework (line 17). After updating the first stage, $N_{initial}$, with a pairwise combination (line 17-18), the search space is pruned. We explore through the remaining space for the top $k$ features and continue looking for other combinations that will improve the overall gain of the framework (line 19).   

In this work, we use color and texture as distingushing semantic features. So, the search space in the above algorithm corresponds to HSV space for color and Gabor space for texture. In the HSV space, we have 8 ranges of H (as shown in Table I) that correspond to the 8 major colors. So, we set $N$=8 and execute Algorithm 1 to select the most optimal individual colors as well as pairwise combinations ($\binom{k}{2}$) for the first stage. For texture, we use the filter-bank approach as discussed in \cite{arivazhagan2003texture, haghighat2013identification}. The initial Gabor space consists of 20 filters \cite{Jain97objectdetection, hu2014optimal} corresponding to 5 scales/frequencies: 4$\sqrt{2}$, 8$\sqrt{2}$, 16$\sqrt{2}$, 32$\sqrt{2}$, 64$\sqrt{2}$ and 4 orientations: 0, 45, 90 and 135 degrees as shown in Fig. 4. So, in case of texture selection, we set $N$=20 and execute Algorithm 1 exploring the individual as well as pairwise combinations of textures to construct the first stage.  We set $k=4 $ in Algorithm 1 during color selection from HSV space and $k=5$ during texture selection from Gabor space. 

\begin{figure}[t]
\centering
\includegraphics[width=0.3\textwidth]{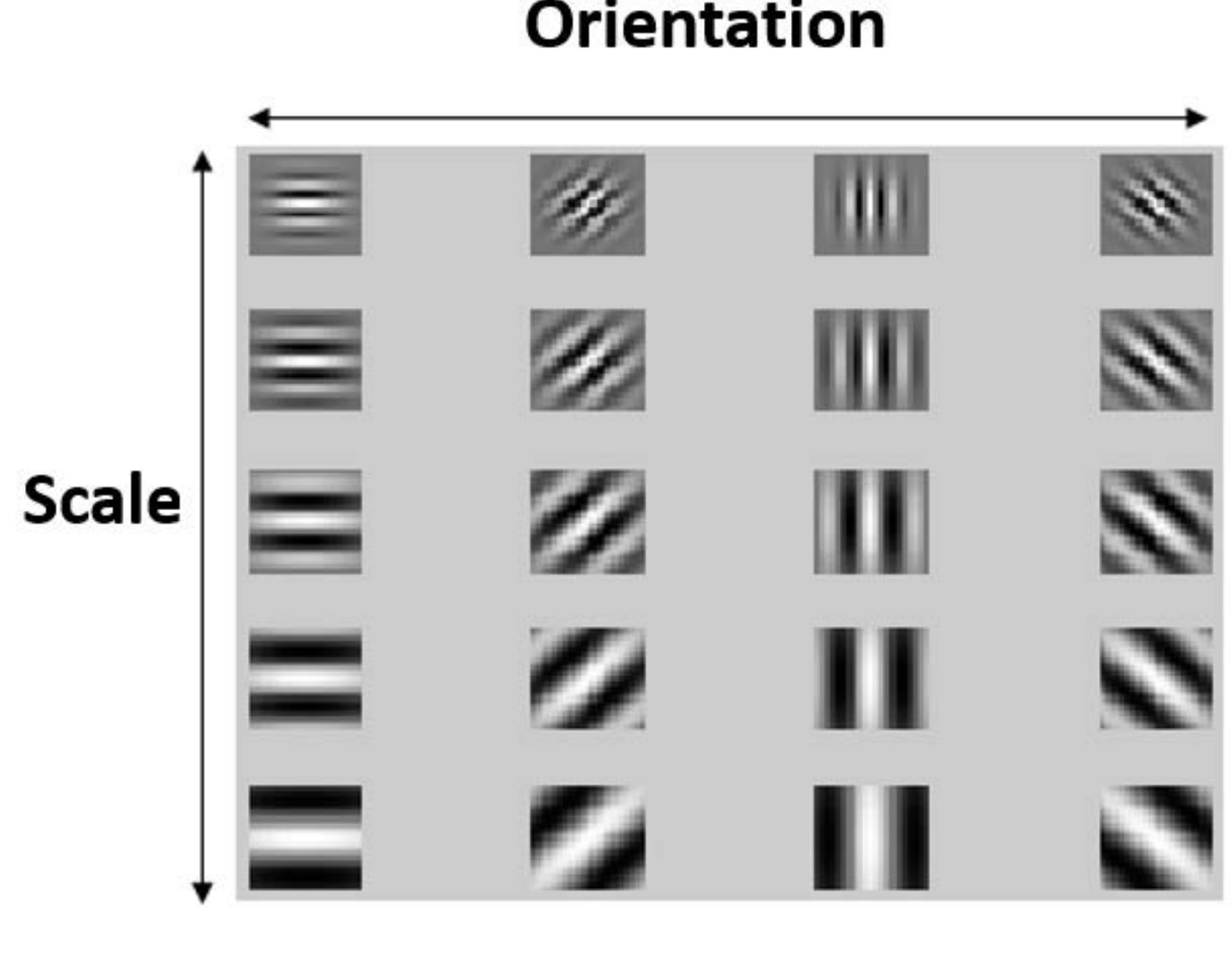}
\caption{Gabor filters corresponding to 5 scales, 4 orientations}
\end{figure}
\section{Design Methodology}
In this section, we describe the procedure for training and testing the hierarchical framework.

\subsection{Training the Hierarchical Framework}

\begin{algorithm}
\caption{Methodology to train the hierarchical framework}
\label{algo2}
 \textbf{Input:} Original classifier $N_{orig}$, training dataset $D_{tr}$ 
 \textbf{Output:} Semantically decomposed hierarchical framework, $N_{hier}$  with optimised first stage
\begin{algorithmic}[1]
\State Call Algorithm 1 on HSV space. initialize the output from Algorithm 1 as  $N_{color}$ = $N_{initial}$ 
\State Call Algorithm 1 on Gabor space. initialize $N_{gabor}$ =  $N_{initial}$ 
\State initialize $N_{combo}= N_{color}$ AND $N_{gabor}$
\State Calculate gain $G_{color}$, $G_{gabor}$, $G_{combo}$  as per Eq. 2 with each $N_{i}$ obtained from Steps 1-3 in the first stage and $N_{orig}$ as second stage.
\State Select the configuration corresponding to max($G_{color}$, $G_{gabor}$, $G_{combor}$) as $N_{initial}$ 
\State $N_{hier}$= $N_{initial}$  with maximum gain as first stage followed by $N_{orig}$ as second stage
\end{algorithmic}
\end{algorithm}
Algorithm 2 shows the pseudocode for training the hierarchical framework. The process is a continuation of the semantic selection algorithm (Algorithm 1) discussed in the previous section. At the end of the semantic selection method, we get the optimal 2-level OR-AND first stage configuration with the best semantics that are representative of the objects of interest in the training dataset. 

Algorithm 2 calls the semantic selection algorithm (Algorithm 1) for color and texture separately and generates the hierarchical framework, $N_{hier}$. Depending upon the efficiency (or gain) values, the algorithm selects either the individual color/texture first stage configuration or selects the ANDed combination of color and texture. As discussed earlier, AND only improves the efficiency of the classifier. By ANDing both texture and color semantic configurations, we are basically filtering out or eliminating more non-objects of interest in the first stage of the hierarchy. However, the color/texture ANDed configuration is selected only when the overall gain due to the improved conditional activation (more inputs filtered at first stage) exceeds the penalty of the additional classifiers in the first stage. As the quality constraint on the individual stages are met from Algorithm 1, it is irrelevant to check for quality (or accuracy) constraint here. The stage that yields the maximum gain or efficiency is chosen as the first stage of the semantically decomposed framework. 

\subsection{Testing the Hierarchical Framework}
\setlength{\textfloatsep}{3pt}
\begin{algorithm}
\caption{Methodology to test the hierarchical framework}
\label{algo3}
 \textbf{Input:} Test instance $I_{test}$, hierarchical framework $N_{hier}$  with appropriate activation threshold \\
 \textbf{Output:} $I_{test}$ classified as clutter or object of interest
\begin{algorithmic}[1]
\State 	Obtain semantic feature vectors of $I_{test}$ which are passed as inputs to the NN(s) comprising the initial stage in $N_{hier}$
\State 	If output of the NN(s) in the initial stage of $N_{hier}$  is such that final classifier or second stage is not enabled, then \textbf{TERMINATE} testing and Output =$I_{test}$  classified as clutter.
\State 	If the initial stage NN(s) produce a sufficient confidence level on the output meeting the OR-AND conditions, then, second stage or final classifier is activated and  Output = Output of final classifier. 
Please note that the input to the final classifier is the original test instance, $I_{test}$  and not the semantic feature vector.
\end{algorithmic}
\end{algorithm}
Algorithm 3 describes the overall testing methodology for the hierarchical framework. Given a test instance, $I_{test}$, the process classifies it as clutter or the object of interest. $N_{hier}$  obtained from the training phase contains appropriate information about the OR-AND operations to define the conditions for which the final stage/classifier will be enabled. $N_{hier}$ also ensures that the proper semantic feature for the test instance is extracted. 

In summary, $N_{hier}$  implicitly extracts the representative semantic information from the input and utilizes the same in the initial stage of the framework thereby ensuring conditional activation of the final stage. Thus, the proposed approach is systematic and hence can be applied to all object detection problems.

\section{Experimental Methodology}
In this section, we describe our experimental setup used to evaluate the performance of the hierarchical framework. We have implemented an ANN based image recognition platform for the Caltech101 dataset \cite{jarrett2009best} which is a large image dataset containing over 30,000 labeled examples of 101 different images. Each classifier used is a feedforward artificial neural network with 3 layers (Input, Hidden and Output). Each of the ANNs are trained using the standard backpropagation algorithm. For up to 50 different images of the dataset, we implemented the hierarchical framework ($N_{hier}$) trained to recognize the particular object of interest from a host of other images (clutter) exploiting both color and texture based semantic information. 
\begin{table}
\vspace{-3mm}
\renewcommand{\arraystretch}{1.3}
\caption{First Stage Configuration of $N_{hier}$ for CALTECH101}
\centering
\vspace{-1em}
\begin{tabular}{c|c|c}
    \hline
    Image  &  Configuration of   & Representations\\
 & first stage  \\
    \hline
    \hline

Soccer Ball & W.B & R:Red;W:White\\
Bonsai	& (Y+R).G & B:Black;Y:Yellow                        \\
Lotus &	R+Y  &  G:Green          \\
Sunflower &	 Y  \\
Stop sign	& R &   \textbf{COLOR}\\
\hline
Brain &	(G1+G2).G3 & G1: (32$\sqrt2$),0$^{\circ}$); G2: (64$\sqrt2$,0$^{\circ}$)\\
Menorah&	G4+G5 &  G3: (32$\sqrt2$,90$^{\circ}$); G4: (32$\sqrt2$,90$^{\circ}$)\\
Revolver&	G6.G7 &  G5: (64$\sqrt2$,45$^{\circ}$); G6: (32$\sqrt2$,0$^{\circ}$)\\
Guitar	&G8+G9 & G7: (64$\sqrt2$,45$^{\circ}$); G8: (16$\sqrt2$,0$^{\circ}$) \\
Starfish&	G10+G11 & G9: (64$\sqrt2$,90$^{\circ}$); G10: (16$\sqrt2$,0$^{\circ}$)\\
& & G11:(64$\sqrt2$,90$^{\circ}$) \textbf{TEXTURE} \\
\hline
\end{tabular}
\vspace{-1mm}
\end{table}
Of the 50 images, the initial stage configurations for 10 different images are shown in Table I. We can see that the first stage is set to different configurations of OR-AND (OR denoted as +, AND denoted as .) by the training methodology described in the previous section for both color and texture. Each of the Gabor filters selected are represented in the table by their corresponding (scale, orientation). The training methodology of the hierarchical framework confirms accuracy or quality check with that of the traditional classifier using OR operation and then it optimizes the efficiency using AND. The Gabor filters/colors selected in the process are also the most optimum semantics for the given set of images. In addition to Caltech101, we evaluated our approach on another dataset CIFAR10 \cite{krizhevsky2010convolutional} which consists of 60,000 colored images belonging to 10 classes. The initial stage configuration for 4 images are shown in Table II.
\begin{table}
\renewcommand{\arraystretch}{1.3}
\caption{First Stage Configuration of $N_{hier}$ for CIFAR10}
\centering
\vspace{-1em}
\begin{tabular}{c|c|c}
    \hline
    Image  &  Configuration of   & Representations\\
 & first stage  \\
    \hline
    \hline

Ship & W+B & R:Red;W:White\\
Truck	& (W+R).B & B:Black   \textbf{COLOR}                      \\
\hline
Airplane &	(G12+G13) & G12: (42$\sqrt2$,22.5$^{\circ}$); G13: (36$\sqrt2$,67.5$^{\circ}$)\\
Toad &	G14.G15 & G14: (10$\sqrt2$,90$^{\circ}$); G15: (28$\sqrt2$,45$^{\circ}$)\\
& &  \textbf{TEXTURE} \\
\hline
\end{tabular}
\vspace{0mm}
\end{table}

In this work, we used software simulations to obtain classification accuracy and hardware simulations to obtain energy values. We implemented the 2-stage semantically decomposed classification framework for the object detection application in Matlab. We measured runtime for the applications using performance counters on Intel Core i7 3.60 GHz processor with 16 GB RAM.
For hardware implementation, we specified each classifier as an accelerator at the register-transfer logic (RTL) level \cite{venkataramani2013quality}. We used the Synopsys Design Compiler to synthesize the integrated design to a 45nm SOI process from IBM. Finally, we used Synopsys Power compiler to estimate energy consumption of the synthesized netlists. We  optimized the logic synthesis for energy using proper timing constraints to ensure that baseline is energy-efficient. 


\section{Results}
In this section, we present the experimental results that demonstrate the benefits of our approach. We use Caltech101 as our primary benchmark to evaluate the benefits with semantic decomposition. 

\subsection{Energy Improvement}
\begin{figure}[t!]
\centering
\includegraphics[width=0.8\linewidth]{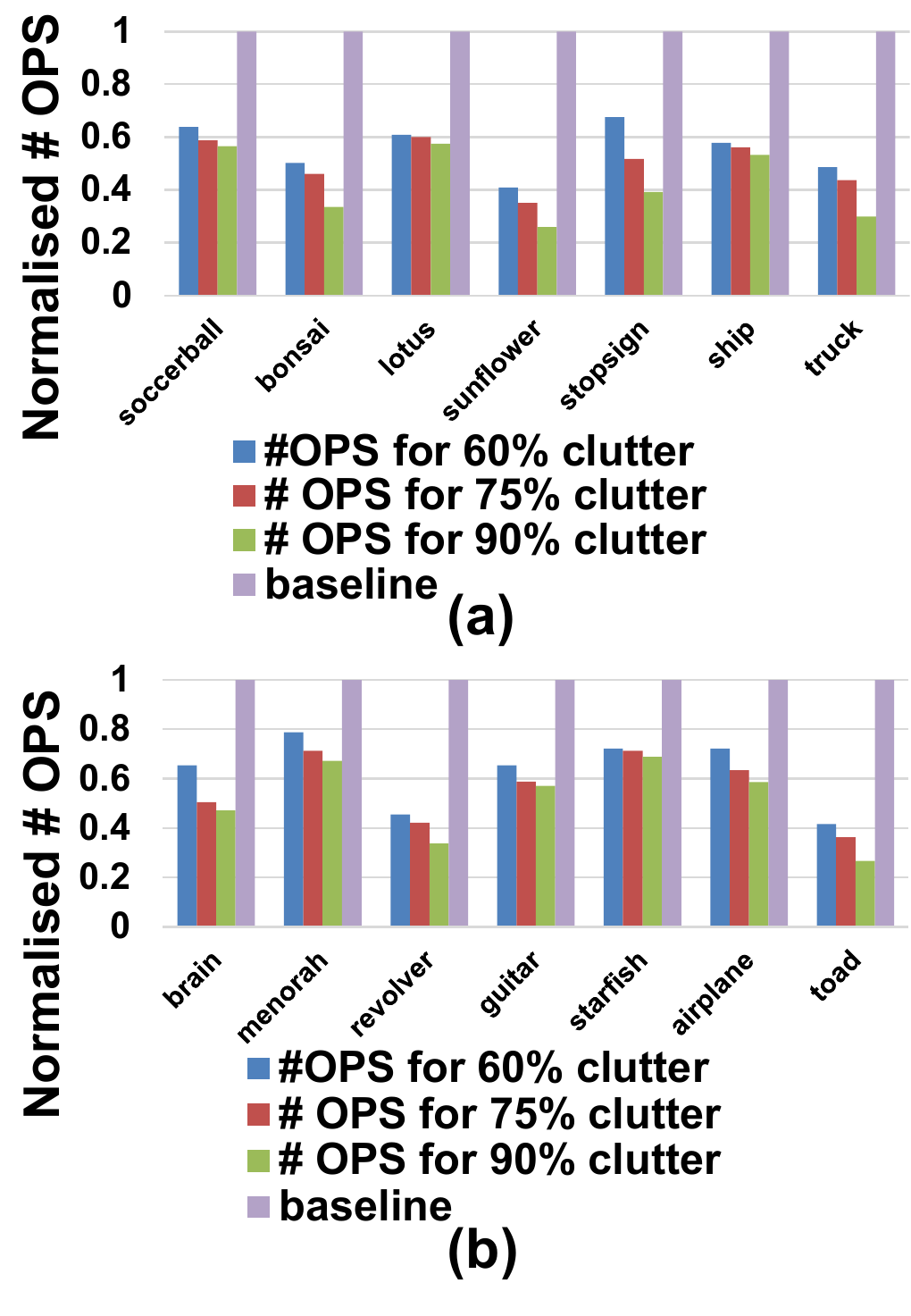}
\vspace{-3mm}
\caption{Normalized OPS for images with (a) color as semantic (b) texture as semantic}
\end{figure}
Fig. 5 (a, b) shows the normalized improvement in efficiency with respect to the single stage classifier (which forms the baseline) for the 14 images of Table I and II. We quantify efficiency as the average number of operations (or total number of MAC computations) per input (OPS).  For each image in the dataset, we varied the fraction of clutter in the test-set (60\%, 75\% and 90\%) and evaluated the efficiency. As mentioned earlier, there is a significant disproportion in the distribution of input data [9]. Thus, in our experiments we evaluated our approach by varying the fraction of clutter(non-objects of interest) in the input data for an object detection task.  We observe that the hierarchical framework provides between 1.97x-2.64x (average: 2.31x) improvement in average OPS/input compared to baseline across the 10 different images for Caltech. For CIFAR, the average reduction in OPS corresponds to 1.88x across the 4 different images. Note that the benefits vary depending on the fraction of clutter in the dataset.

Fig. 5 clearly illustrates that maximum benefit for each image is observed when the fraction of clutter is 90\%. This can be corroborated to the fact that the initial stage filters out a lot of the object of interest. In case of hardware implementation, the reduction in OPS for Caltech translates on an average to 1.64x-2.05x (average: 1.93x) improvement in energy with variation of clutter as illustrated in Fig. 6. For CIFAR10, the average improvement in energy with variation of clutter is 1.46x. For software implementation, the reduction in OPS converts to an average of 2.23x/1.95x improvement in runtime for Caltech/CIFAR10 respectively as shown in Fig. 7.
\begin{figure}[t!]
\centering
\includegraphics[width=0.8\linewidth]{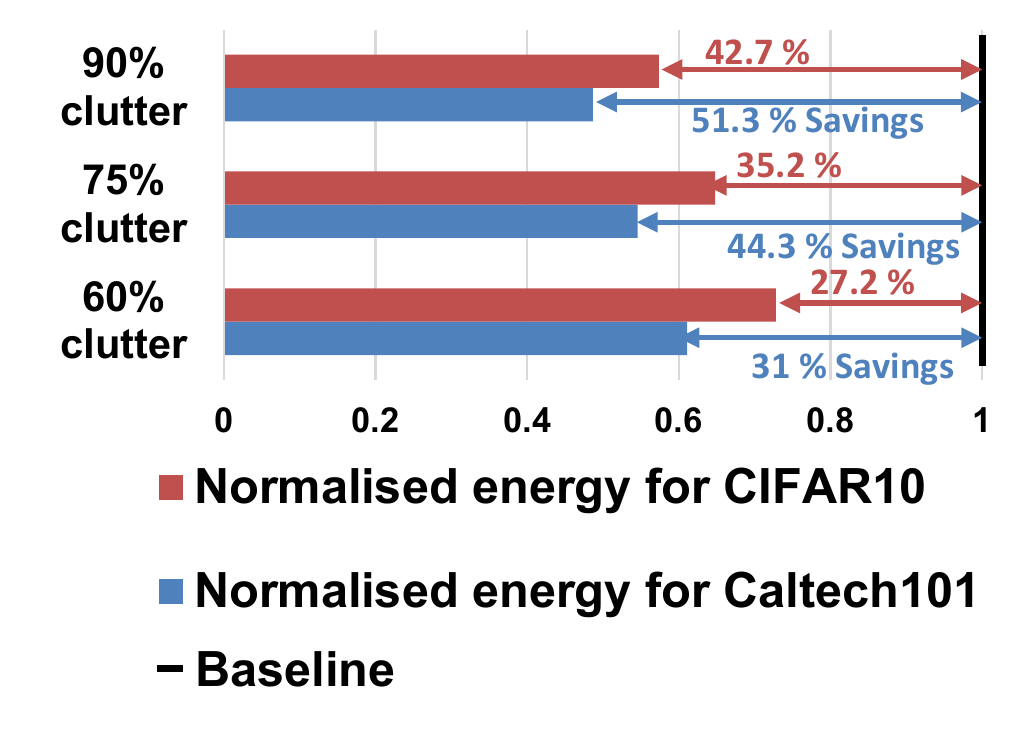}
\vspace{-3mm}
\caption{Average hardware energy for different fraction of clutter in the dataset}
\vspace{-2mm}
\end{figure}

\begin{figure}[t!]
\centering
\includegraphics[width=0.8\linewidth]{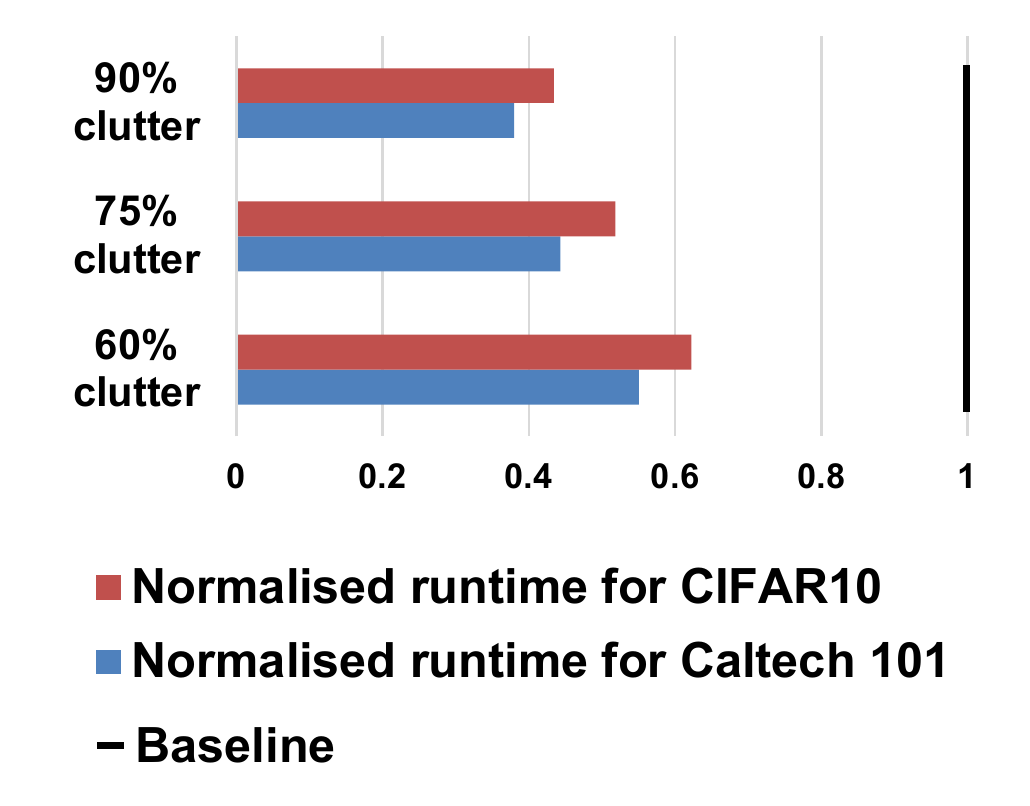}
\vspace{-3mm}
\caption{Average software energy/runtime for different fraction of clutter in the dataset}
\end{figure}

\subsection{Impact of Variation of Clutter on Efficiency}
Here, we examine the impact of clutter on the overall efficiency of our proposed hierarchical framework. It is evident that the main idea behind semantic decomposition and building the hierarchical framework is to reject majority of clutter images in the initial stage. Being less complex than the final, the initial stage would contribute less to the overall computational cost. So, the final stage should ideally get enabled only for the objects of interest and no clutter data at all. However, in practicality, the clutter data will have certain semantic information representative of the object of interest and will thus be passed to the next stage. Fig. 8 shows the fraction of clutter that is actually passed to the final stage as the clutter percentage is varied gradually from 60\% to 90\% for both CALTECH and CIFAR. We observe that as the clutter fraction is increased, the amount of clutter filtered out increases that correspond to lesser number of activation of the final stage. For instance, for 60\% clutter images in the CIFAR10 dataset, the final stage is activated for 45.5\% clutter, while for 90\% clutter it is activated for 33.8\%. Thus, we observe maximum savings in both energy and OPS as the fraction of clutter increases (Fig. 5, 6, 7).

\begin{figure}[t!]
\centering
\includegraphics[width=0.8\linewidth]{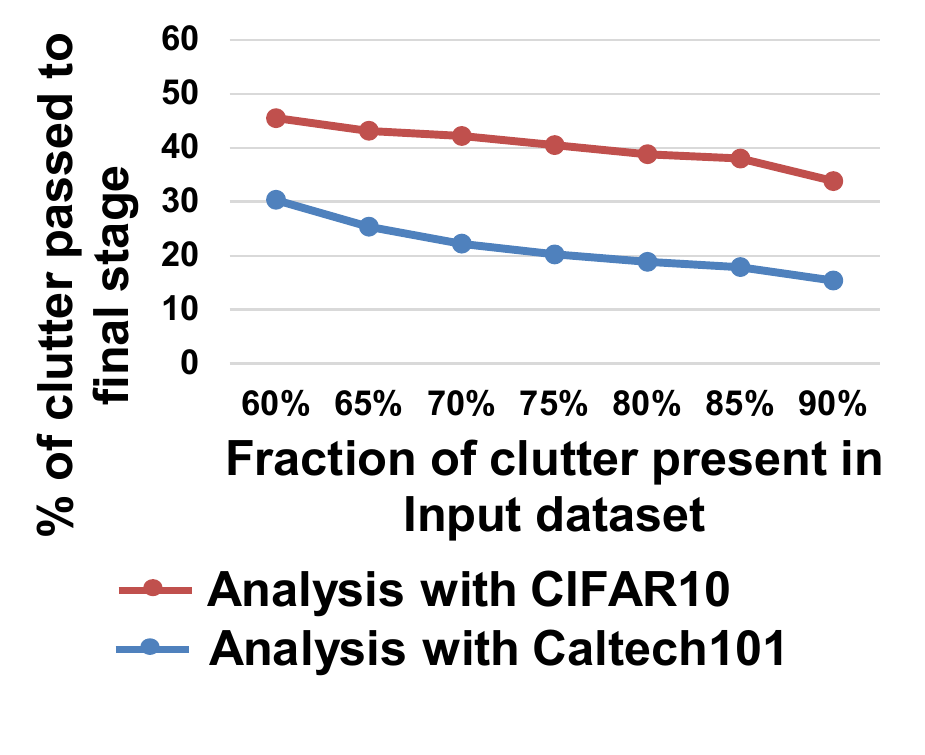}
\vspace{-8mm}
\caption{Fraction of clutter passed to final stage of $N_{hier}$  as clutter \% is increased in the input dataset}
\end{figure}

\subsection{Optimizing the complexity of the first stage}
The hierarchical design methodology during training first  meets the output quality or accuracy constraint and then optimizes the framework to get maximum efficiency. In order to get the most benefits, we need to filter out more clutter in the initial stage. We can achieve this by increasing the complexity of the first stage by adding more neurons to the hidden layer. Fig. 9 shows the normalized energy of the entire hierarchical framework as the complexity of the first stage (Y from Table I) is varied for detecting lotus from the Caltech101 dataset. It can be clearly seen that the amount of clutter filtered increases with the increasing complexity of the first stage. So, as the initial stage becomes more complex, the final stage is enabled for fewer clutter data from the total fraction of clutter. Thus, in the beginning, we observe a decreasing trend in energy. However, the increasing complexity of the first stage would also add an additional overhead to the cost computation that would at some point overcome the total cost savings. This break-even point corresponds to the maximum benefits or the lowest energy that we can achieve using the hierarchical framework for this particular example. Beyond this point, the cost increases. In Fig. 9, we see that the break-even point corresponds to 0.508 (Normalized energy)  which translates to 1.97x improvement in computational cost. This behavior is taken into account in our design methodology described in the previous section.
\begin{figure}[t!]
\centering
\includegraphics[width=0.9\linewidth]{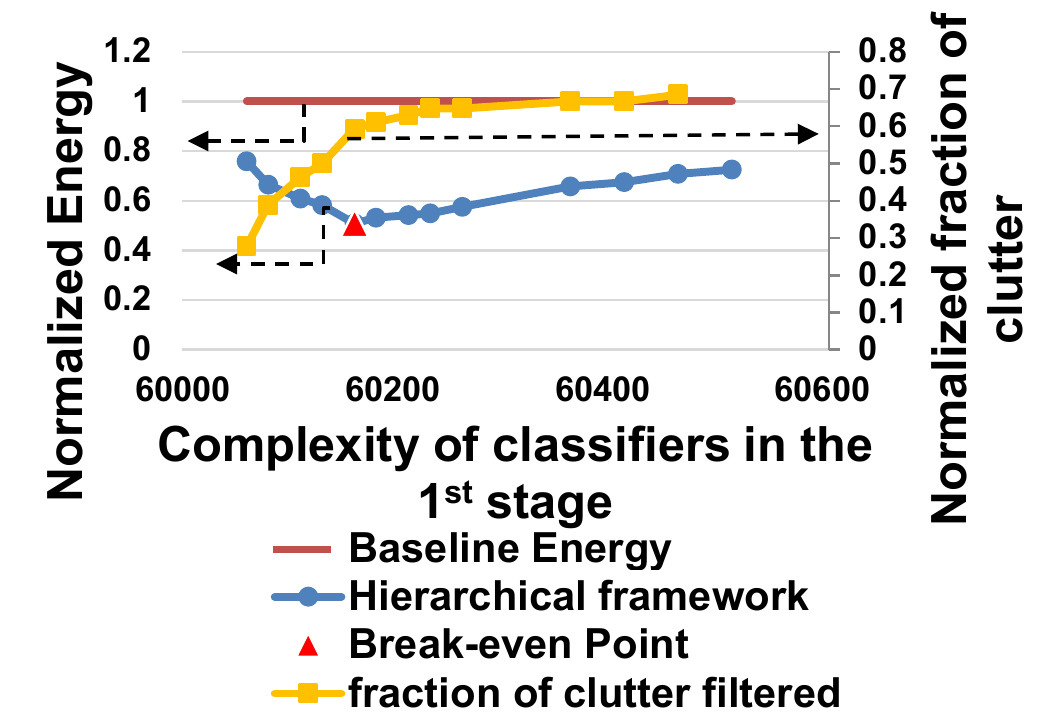}
\vspace{-3mm}
\caption{Normalized reduction in energy by varying complexity of 1st stage (Y) in $N_{hier}$ for detecting lotus from Caltech101 dataset}
\end{figure}

\subsection{Efficiency-Accuracy Tradeoff using Confidence level (\textbf{$\delta$})}
In Section III, 
we discussed that the confidence level or activation threshold ($\delta$) can be regulated to modulate the amount of clutter being passed to the final classifier and further optimize the efficiency. Fig. 10 shows the normalized energy of the hierarchical framework as the accuracy of the first stage is varied by changing the $\delta$ value for the Caltech 101 dataset. The tradeoff analysis helps us attain the most optimum $\delta$ of a semantically decomposed framework for an object detection task, that yields maximum computational benefits while maintaining comparable accuracy with that of the traditional single stage classifier.

Setting $\delta$ to a low value implies that more clutter will now be misclassified by the first stage, and forwarded to the final classifier. Increasing $\delta$ would result in lesser clutter being misclassified thus improving the overall accuracy of the first stage as can be seen from Fig. 10. 
Beyond a particular $\delta$, the objects of interest will be misclassified and filtered out. This $\delta$ value corresponds to the maximum overall accuracy of the first stage in the hierarchy. In Fig. 10, we observe that as the normalized accuracy value increases from 0.86 ($\delta$=0.1) to 0.95 ($\delta$=0.4), there is a 2.25x reduction in total \# OPS which quantifies energy-efficiency. In this case, beyond $\delta$=0.4 the accuracy declines and hence those $\delta$ values are not considered. Please note that, the energy benefits will continue to increase beyond $\delta$ = 0.4 as the second stage of the hierarchy is enabled for less instances with increasing $\delta$. Thus, for applications that permit tolerable accuracy loss, we can use higher values of $\delta$  to get higher energy benefits. Please note that the accuracies shown in Fig. 10 are normalized with respect to the baseline accuracy (\~97.8\% in this case). 

 For our OPS evaluations with Caltech 101 and CIFAR10 (Fig. 5), we use $\delta$ =0.4 and 0.55 respectively as obtained from the tradeoff analysis.   

\begin{figure}[t!]
\centering
\includegraphics[width=0.8\linewidth]{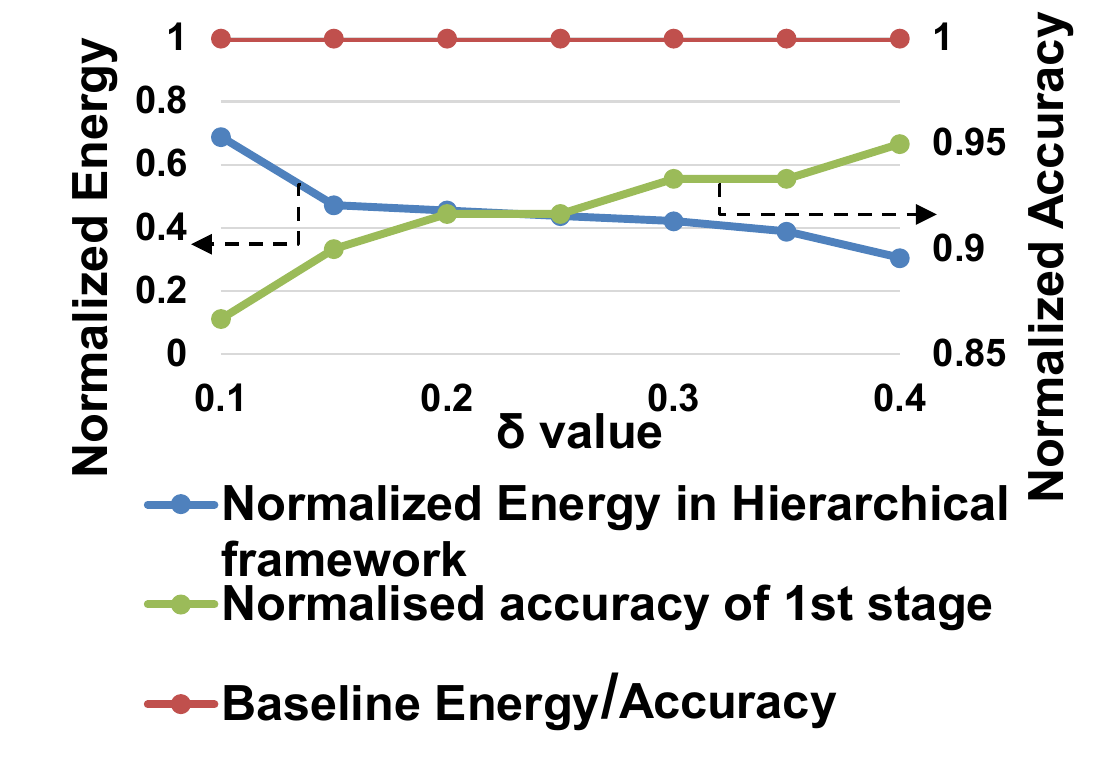}
\vspace{-5mm}
\caption{Normalized energy vs. accuracy tradeoff by modulating confidence level ($\delta$)}
\end{figure}

\subsection{Impact of addition of first stage to the overall training time}
Until now, the energy benefits observed correspond to the reduction in testing complexity for a given task using our proposed hierarchical framework. However, the first stage also adds an additional overhead on the total training time to construct the 2-stage framework. Fig. 11 shows the overall normalized training time of the hierarchical framework for detecting the 10 images of Table I (Caltech 101). %
The first stage training time shown includes the additional time expended during the iterative optimal semantic selection process (\textit{Algorithm 1} in Section III). It is clearly seen that the overall training time of the hierarchical framework is greater than the baseline single stage classifier for each image. We also observe that the first stage construction with color features (Fig. 11 (a)) takes lesser time than that of texture (Fig. 11 (b)). This can be attributed to two factors: a) Gabor filtering is computationally more expensive involving complex operations than HSV. b) In \textit{Algorithm 1}, for color features, we only explore the 8 feature search space to select the optimal color in comparison to 20 features for texture. The increased search space further adds to the training time. On an average, we observe that the training time increases by 18.4\% in Fig. 11 (a) and 31.4\% in Fig. 11 (b) compared to the baseline traditional classifier. While there is a training overhead with our proposed framework, in typical object detection applications, training is performed only once or very infrequently. Testing, on the other hand, is done more frequently over longer periods of time. Since our proposed framework yields significant reduction in testing energy, a small increase in training cost is a favorable tradeoff. 

\begin{figure}[t!]
\centering
\includegraphics[width=0.8\linewidth]{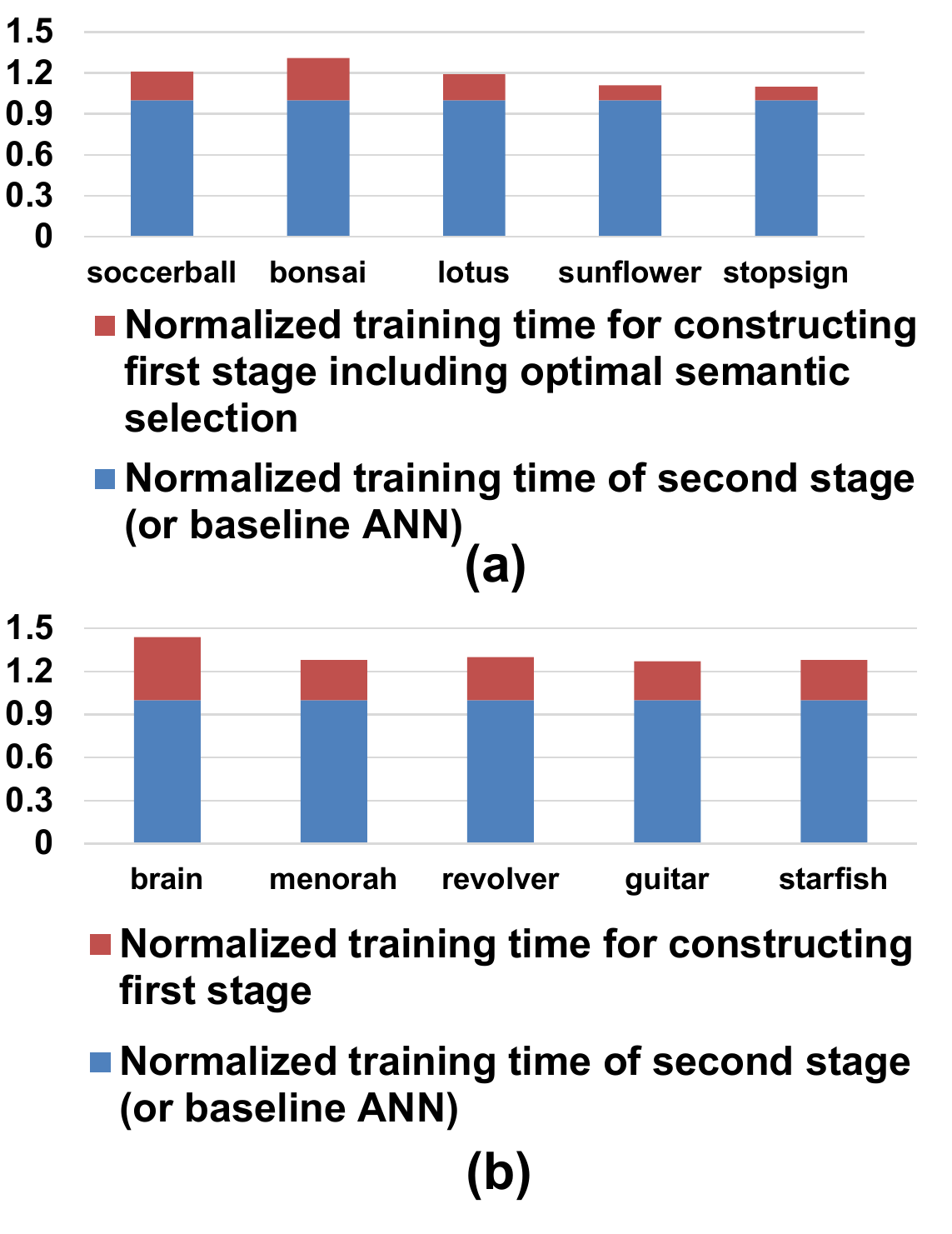}
\vspace{-5mm}
\caption{Normalized training time for images with (a) color as semantic (b) texture as semantic}
\end{figure}

\subsection{Combining Color and Texture in the Initial Stage}

For a given dataset, the hierarchical training methodology first constructs the individual color/texture configurations. Then, the individual color/texture stages are ANDed together. In case, the overall gain of the hierarchical framework improves with the ANDed configuration, the color(AND)texture combination is selected as the first stage of the hierarchy. For certain images in the Caltech101 dataset shown in Table III, such configuration was chosen. For all the images, the semantic selection methodology chose the corresponding colors and textures individually for initial stage construction. The training methodology, in addition to color ANDed the appropriate Gabor features in the initial stage. In order to observe the additional benefits with the ANDed configuration and for comparison purpose, we implemented a separate semantically decomposed framework using only color configuration obtained from Algorithm 1. 
\begin{table}
\renewcommand{\arraystretch}{1.3}
\caption{First Stage Configuration}
\centering
\vspace{-1em}
\begin{tabular}{c|c|c}
    \hline
    Image  &  Configuration with & Configuration with
\\ & color \& texture  & color only\\
    \hline
    \hline

    Motor bike   &   (G1+G2).B & B\\
    \hline
 Buddha  &   (G3+G4).Y & Y\\
   \hline
Flamingo   &   (G5+G6).R & R\\
   \hline  
\end{tabular}
\vspace{-3mm}
\end{table}

We conducted a similar set of experiments on the Caltech101 dataset by varying the fraction of clutter on the combined color/texture configuration. Fig. 12 shows the normalized \#OPS for the images in Table III (Representations are same as Table I). We observe that the framework provides 1.35x-1.79x improvement (with respect to the baseline) in average OPS per input as the clutter fraction is increased from 60\% to 90\%. Note maximum benefits correspond to larger clutter fraction in the dataset. It is evident that the benefits observed are due to conditional activation of the final stage. Now, in the combined configuration, the initial stage consists of AND operation. Hence, we can deduce that the final classifier in this case would get activated for lesser number of instances as compared to the configuration with only color as semantic. Fig. 13 shows the average energy in both cases (color, color AND texture) as the clutter fraction is varied. It is clearly seen that color AND texture configuration gives more savings than the latter. This is due to the fact that the benefits of reduced final stage activation in the combined case overcomes the penalty due to the addition of texture configuration in the first stage. Thus, our proposed design methodology ensures maximum cost savings by selecting the most optimum semantic configuration.
\begin{figure}[t!]
\centering
\includegraphics[width=0.8\linewidth]{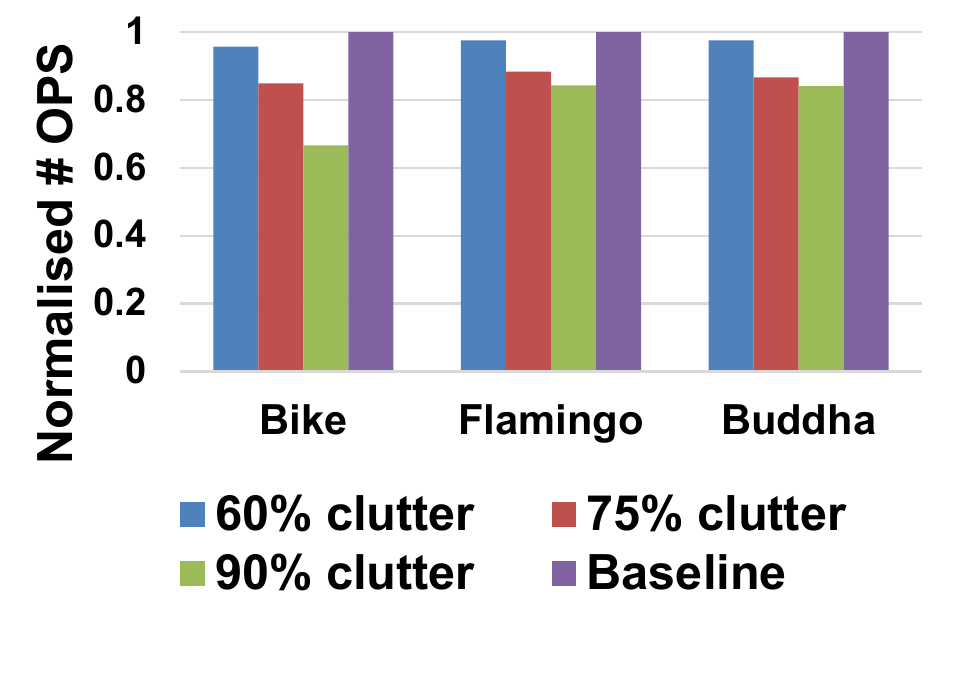}
\vspace{-4mm}
\caption{Normalized OPS for images with color and texture as semantic}
\vspace{-1mm}
\end{figure}
\vspace{-2mm}
\begin{figure}[t!]
\centering
\includegraphics[width=0.8\linewidth]{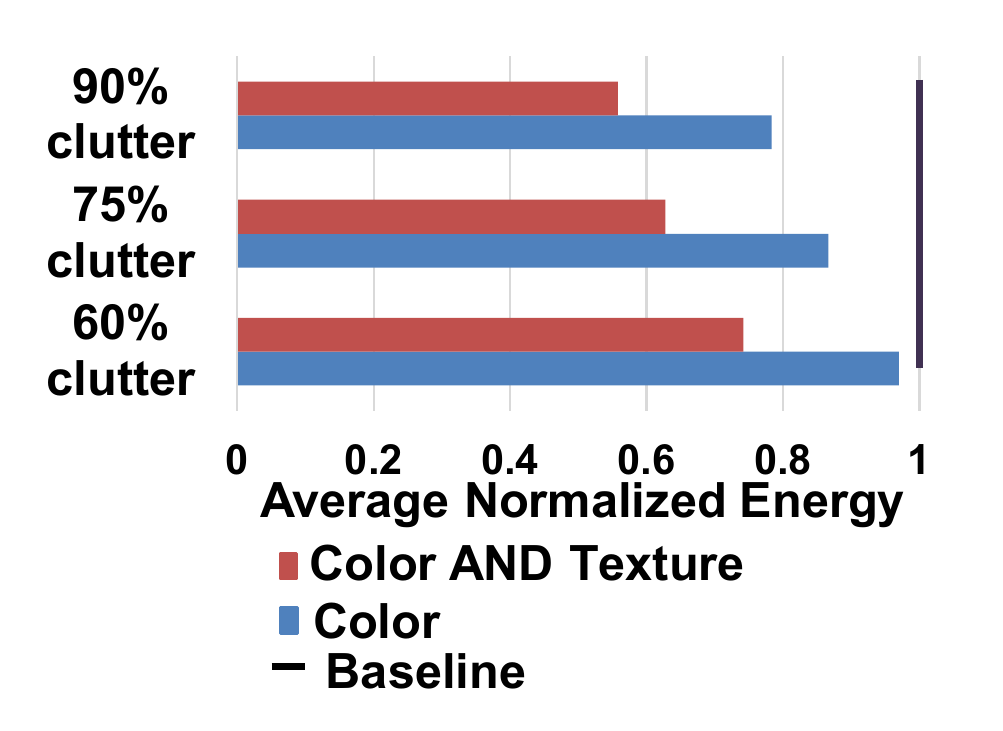}
\vspace{-5mm}
\caption{Average normalized energy for both configurations: color only and combined color/texture}
\vspace{1mm}
\end{figure}
\section{Conclusion}
We presented a systematic approach to optimize energy-efficiency of machine learning classifiers by exploiting the characteristic semantic information of inputs. We observe that certain semantic information is common across various objects in a dataset. We use the common semantic features to distinguish the objects of interest from the remaining inputs in object detection applications. Based on the above insight, we proposed the concept of hiererchical classification based on semantic decomposition. We develop a systematic methodology to implement a 2-stage semantically decomposed classification framework using color/texture as semantic information. We achieve this by arranging the classifiers (ANNs) in increasing order of complexity as per the characteristic semantic features they are trained to recognise. The design methodology is equipped to implicitly gather the most appropriate semantic information for optimum efficiency. To quantify the potential of semantic decomposition, we used color and texture as a basis for segmentation and designed the hierarchical framework for object detection for various images of the Caltech101/CIFAR10 dataset. Color and texture information were obtained using HSV and Gabor filtering operations, respectively. Our experiments demonstrate significant improvement in energy over hardware implementation with respect to traditional approach.

\section*{Acknowledgment}
This work was supported in part by C-SPIN, one of the six centers of StarNet, a Semiconductor Research Corporation Program, sponsored by MARCO and DARPA, by the Semiconductor Research Corporation, the National Science Foundation, Intel Corporation and by the National Security Science and Engineering Faculty Fellowship.

\ifCLASSOPTIONcaptionsoff
  \newpage
\fi

\end{document}